\documentclass{article}

\usepackage{arxiv}

\usepackage[utf8]{inputenc} 
\usepackage[T1]{fontenc}    
\usepackage{hyperref}       
\usepackage{url}            
\usepackage{booktabs}       
\usepackage{amsfonts}       
\usepackage{nicefrac}       
\usepackage{microtype}      
\usepackage{lipsum}		
\usepackage{graphicx}
\usepackage{natbib}
\usepackage{doi}
\usepackage{CJKutf8}
\usepackage{color}
\usepackage{enumitem}
\usepackage{tabularx}
\usepackage{graphicx}
\usepackage{multirow}
\usepackage{bbding}

\usepackage[marginal]{footmisc}

\title{DuTrust: A Sentiment Analysis Dataset for \\ Trustworthiness Evaluation}

\author{ Lijie Wang$^\dag$, Hao Liu$^\dag$, Shuyuan Peng, Hongxuan Tang, Xinyan Xiao, Ying Chen, Hua Wu, Haifeng Wang \\
	Baidu Inc., Beijing, China \\
	\texttt{\{wanglijie,liuhao24,pengshuyuan,tanghongxuan,xiaoxinyan,chenying04,hua\_wu,wanghaifeng\}@baidu.com} \\
}

\renewcommand{\shorttitle}{\textit{arXiv} Template}

\hypersetup{
pdftitle={A template for the arxiv style},
pdfsubject={q-bio.NC, q-bio.QM},
pdfauthor={David S.~Hippocampus, Elias D.~Striatum},
pdfkeywords={First keyword, Second keyword, More},
}

\begin{document}
\maketitle
\footnote{$\dag$ They contribute equally to this work.\\}
\begin{abstract}

While deep learning models have greatly improved the performance of most artificial intelligence tasks, they are often criticized to be untrustworthy due to the black-box problem. 
Consequently, many works have been proposed to study the trustworthiness of deep learning. However, as most open datasets are designed for evaluating the accuracy of model outputs, there is still a lack of appropriate datasets for evaluating the inner workings of neural networks. The lack of datasets obviously hinders the development of trustworthiness research.
Therefore, in order to systematically evaluate the factors for building trustworthy systems, we propose a novel and well-annotated sentiment analysis dataset to evaluate robustness and interpretability. To evaluate these factors, our dataset contains diverse annotations about the challenging distribution of instances, manual adversarial instances and sentiment explanations.
Several evaluation metrics are further proposed for interpretability and robustness.
Based on the dataset and metrics, we conduct comprehensive comparisons for the trustworthiness of three typical models, and also study the relations between accuracy, robustness and interpretability.
We release this trustworthiness evaluation dataset at \url{https://github/xyz} and hope our work can facilitate the progress on building more trustworthy systems for real-world applications.
\end{abstract}


\section{Introduction}
\label{sec:intro}

In the last decade, deep learning has been rapidly developed and has greatly improved various artificial intelligence (AI) tasks in terms of accuracy (\cite{deng2014deep,litjens2017survey,pouyanfar2018survey}), such as Natural Language Processing (NLP) and Computer Vision (CV). However, as deep learning models are black-box systems, their inner decision processes are opaque to users. This lack of transparency makes them untrustworthy and hard to be applied in the decision-making applications, such as health, commerce and law (\cite{fort2016yes}), where users often hope to understand the decision-making process of the output. Consequently, there is a growing interest in designing trustworthy deep learning systems. The studies of trustworthiness mainly focus on two properties: interpretability (\cite{simonyan2014deep,ribeiro2016should,deyoung2020eraser}) and robustness (\cite{ribeiro2020beyond,gui2021textflint}). 

In the line of \textit{interpretability}, researchers aim to provide a human with explanations in understandable terms (\cite{doshi2017towards, zhang2020survey}). 
Two-model methods use independent extraction and prediction modules, so as to first extract explanations from the input and then make predictions based on the extracted explanations (\cite{lei2016rationalizing, jain2020learning}).
Saliency methods regard the well-trained model as a black-box and use gradient or attention to extract explanations to reveal the reasons of prediction (\cite{simonyan2014deep, sundararajan2017axiomatic, smilkov2017smoothgrad}).
Interpretability evaluation is also studied (\cite{belinkov2019analysis, jacovi2020towards, deyoung2020eraser, ding2021evaluating}) from the perspectives of plausibility and faithfulness.

Another line of research studies \textit{robustness}, which enhances the deep learning models to tolerate the adversarial attacks and imperceptible additive noises (\cite{heaven2019deep}). Various adversarial techniques for crafting adversarial examples have been proposed for different NLP tasks (\cite{alzantot2018generating, rychalska2019models, ribeiro2020beyond, gui2021textflint}). 
For example, interactive annotation methods ask non-expert annotators to construct adversarial examples which can expose model brittleness (\cite{ettinger2017towards, nie2020adversarial}); and pattern-based methods exploit spurious patterns summarized from the dataset to generate adversarial examples and use them to evaluate models (\cite{glockner2018breaking, mccoy2019right, ribeiro2020beyond, gui2021textflint}).

\begin{figure}[tb]
\centering
\includegraphics[width=0.98\textwidth]{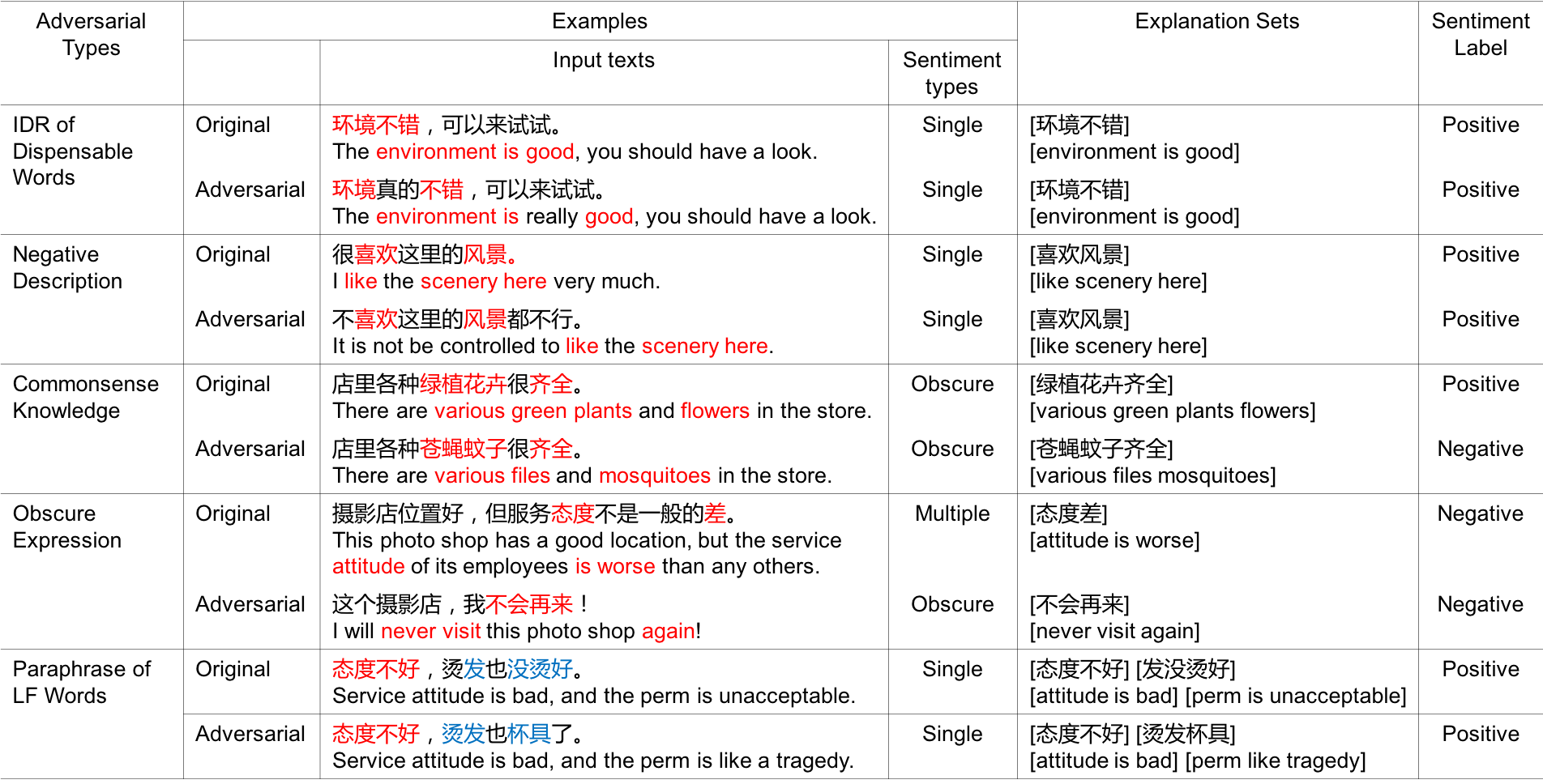}
\caption{Examples in our evaluation dataset. The first column shows the perturbation types of adversarial instances. The forth column labels the types of sentiment expression, and the fifth column lists the manually annotated token-level based explanations. An input text may have more than one explanation set, and tokens in the same [] belong to the same explanation set. please see Section \ref{sec:data} for details.}
\label{fig:intro}
\end{figure}

\begin{table*}[tb]
\renewcommand\tabcolsep{2.5pt}
\centering
\begin{tabular}{l | c | c | c}
\toprule
\multirow{2}{*}{Evaluation Datasets} & \multirow{2}{*}{Size} & \textbf{Robustness} & \textbf{Interpretability} \\
& & Adversarial generation method & Explanations annotation level\\
\hline
CHECKLIST (\cite{ribeiro2020beyond}) & - & pattern-based generation & \XSolidBrush \\
TextFlint (\cite{gui2021textflint}) & - & pattern-based generation & \XSolidBrush \\
BuildItBreakIt (\cite{ettinger2017towards}) & - & interactive manual annotation & \XSolidBrush \\
ERASER (\cite{deyoung2020eraser}) & 2,000 & \XSolidBrush & snippet-level annotation \\
\hline
Ours & 3,004 & manual annotation & token-level annotation \\
\bottomrule
\end{tabular}
\caption{The Statistics of existing evaluation dataset for trustworthiness.}
\label{tab:dataset_history}
\end{table*} 

Despite these extensive works are already proposed, the evaluation of trustworthiness remains a big open challenge, due to the lack of comprehensively annotated datasets. Table \ref{tab:dataset_history} shows the statistics of existing evaluation datasets from the perspectives of robustness and interpretability. There are two main deficiencies. Firstly, there is still a lack of appropriate manually labeled dataset either for robustness or interpretability. The robustness evaluation often relies on automatically constructed adversarial examples based on linguistic patterns. Such pattern-based generated examples are quite different with real world data, and would be hard to correctly evaluate the robustness of the model in real applications as shown in our experiment. 
To the best of our knowledge, ERASER (\cite{deyoung2020eraser}) is the only dataset available for interpretability evaluation. However ERASER only gives snippet-level explanations, which are too coarse to accurately evaluate the process of prediction. Secondly, and most importantly, interpretability and robustness are often studied and evaluated separately, despite they are correlated factors for trustworthiness. This make it difficult to systematically evaluate the factors of trustworthiness and discover the hidden relationship between these factors.

In order to address the above problems, we release a new and well-annotated dataset for trustworthiness evaluation. We systematically annotated three types of information for sentiment analysis (see Figure \ref{fig:intro}), including challenging types of sentiment, manually-generated adversarial examples and manually annotated explanations. This data allows us to study the trustworthiness of deep learning models and the relationship between accuracy, interpretability, and robustness. Our contributions include: 
\begin{itemize}[leftmargin=*]
\item To the best of our knowledge, this is the first work of trustworthy evaluation dataset. Our data can be used to comprehensively evaluate the trustworthiness of a model from the perspectives of both robustness and interpretability. 
\item We propose several metrics to evaluate model trustworthiness from diverse aspects, including faithfulness, plausibility and robustness.
\item We implement four models with different architectures and parameter sizes, and report their performance based on three interpretation methods. We find that all models don't perform well on the manually annotated adversarial instances and explanations, and no single model can achieve the best performance on all metrics. We believe these findings would help future work to build trustworthy systems.
\end{itemize}

\section{Background}
\label{sec:related-work}
There are diverse research works for trustworthiness in NLP. Various approaches are designed to  verify the trustworthiness from different aspects, which mainly include robustness and interpretability. \textit{Robustness} often means that a model should be able to tolerate adversarial attacks and noisy inputs. \textit{Interpretability} aims to provide explanations of the inner prediction process in a human understandable way.

As a background, we will introduce trustworthiness works about dataset and evaluation in the following paragraphs. Firstly, we introduce works on robustness evaluation, most of which construct various challenging examples to test the model. Then we introduce related works on interpretability evaluation criteria. As the interpretation metrics are used to evaluate the quality of given explanations, we also introduce works about post-hoc interpretation methods used to extract explanations from input.

\paragraph{Robustness Evaluation} Recently, many researches aim to test the robustness of NLP models by creating challenging datasets.
\citet{alzantot2018generating} test well-trained models on adversarial examples which are generated via word replacement and have similar semantics and syntax with the original ones. Then they show that the well-trained sentiment analysis and textual entailment models are fooled by these generated examples with success rates of 97\% and 70\%, respectively. 
\citet{rychalska2019models} evaluate the robustness of models via testing model stability in a natural setting where text corruptions such as keyboard errors or misspelling occur, and find that even strong models fail to achieve sufficient robustness. 
\citet{mccoy2019right} summarize three fallible structural heuristics which are likely learned by the models, and design a new evaluation dataset to illuminate interpretable shortcomings in state-of-the-art (SOTA) models. 
While previous approaches focus either on individual tasks or special behaviors, \citet{ribeiro2020beyond} propose CHECKLIST, a task-agnostic methodology to test comprehensive behaviors of NLP models. CHECKLIST provides a list of linguistic capabilities which are applicable to most tasks, and then use them to create various evaluation examples.
Similarly, \citet{gui2021textflint} propose a multilingual robustness evaluation platform TextFlint for NLP tasks that incorporates universal and task-specific adversarial techniques to evaluate the robustness of the model. 

\paragraph{Interpretability Evaluation} Many works study the definition and evaluation of interpretability. 
\citet{herman2017promise} point out that faithfulness, plausibility, and readability are popular standards of an explanation's quality. They propose a distinction between explanations that are best describe the underlying model and explanations that are tailored to the end user.
\citet{jacovi2020towards} emphasize the difference between plausibility and faithfulness, and provide an explicit form to define faithfulness. In addition, they call for evaluating faithfulness on a continuous scale of acceptability rather than viewing it as a binary proposition.
\citet{deyoung2020eraser} propose the ERASER benchmark which comprises multiple datasets and their corresponding snippet-based explanations to evaluate the interpretability of NLP models. And they evaluate the interpretability from the perspectives of plausibility and faithfulness.
According to recent studies, there are two particularly notable criteria for interpretability: plausibility and faithfulness. However, due to the lack of annotated explanations, there are little studies on the evaluation of interpretation plausibility.

\paragraph{Post-hoc Interpretation Methods} Post-hoc interpretation methods seek to explain why a model makes a specific prediction for a given input. At a very high level, these methods assign an important score to each token in the input.
There are mainly three types of interpretations: gradient-based saliency maps (\cite{simonyan2014deep, smilkov2017smoothgrad, sundararajan2017axiomatic}), linear-based local explanation methods (\cite{ribeiro2016should}), and attention-based methods (\cite{jain2019attention, wiegreffe2019attention, pruthi2020learning}). 
Gradient-based methods determine token importance using the gradient of the loss with respect to the tokens (\cite{simonyan2014deep}). \citet{sundararajan2017axiomatic} introduce integrated gradients, where token importance is determined by integrating the gradient along the path from a baseline input to the original input. \citet{smilkov2017smoothgrad} introduce SmoothGrad by adding small noises to each token embedding and averaging the gradient value over noises.
\citet{ribeiro2016should} propose an explanation technique called LIME to explain the predictions of any classifier by approximating it locally with an traditional linear model.
Attention-based methods use attention scores as token important scores. \citet{wiegreffe2019attention} propose four alternative tests to determine when/whether attention scores can be used as explanations and prove the usefulness of attention mechanisms for interpretability. In this paper, we implement the above methods and compare them in experiments, as shown in section \ref{sec:result}.

Overall, despite there are already some works for trustworthiness evaluation, there remains much to be done. Firstly, there is still a lack of appropriate manually labeled dataset either for robustness or interpretability. As shown before, robustness evaluation often relies on automatically generated adversarial examples, which have not occurred in real applications. Thus, it is hard to expect such an evaluation to be useful for real-world applications. For interpretability, only coarse snippet-level explanations (\cite{deyoung2020eraser}) are available, which fail to give an accurate evaluation for interpretability. Secondly, and most importantly, interpretability and robustness are often studied and evaluated separately, despite they are close properties for trustworthiness. This make it difficult to comprehensively evaluate the trustworthiness and discover the hidden relationship between these capabilities.

\section{Evaluation Dataset Construction}
\label{sec:data}
\begin{figure}[tb]
\centering
\includegraphics[width=0.80\textwidth]{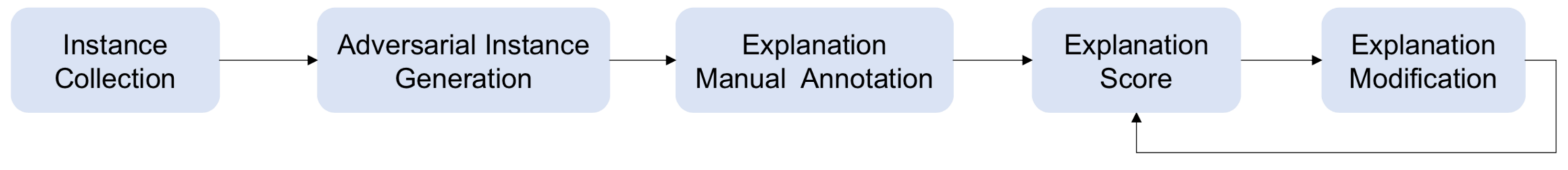}
\caption{The construction workflow of our evaluation dataset.}
\label{fig:data_build}
\end{figure}

In order to appropriately evaluate model trustworthiness, we construct a manually annotated sentiment analysis dataset. This dataset can be used to evaluate specific factors of trustworthiness, including robustness and interpretability, and can be used to study the relations between these factors. We believe this data can help researchers to systematically study model trustworthiness.

Firstly, we create the challenging distribution of instances from practical applications. We classify real-world instances into three types according to the lexical cues of sentiment, including single-sentiment, multiple-sentiment and obscure-sentiment. Although the latter two types which are harder to predict are less happened in real-world applications, we create an equal number of instances for each type, and get a much more challenging distribution of instances. Then, for each instance, we ask annotators to label sentiments and manually create adversarial examples based on common disturbance types of mistakable cases in real applications. Finally, explanations are annotated for all original and adversarial instances.

\subsection{Instance Collection}
\label{ssec:instance_collection}
In order to collect instances from real users, we randomly sample 60,000 instances from the logs of an open sentiment analysis API with the permission of users.
We classify instances into three types: single-sentiment, multiple-sentiment and obscure-sentiment, based on the analysis of the collected instances. As shown in the forth column of Figure \ref{fig:intro}, these types represent different difficulties of prediction, and can reveal the weakness of evaluated models. Obviously, multiple-sentiment and obscure-sentiment instances are more difficult to be predicted. Despite these two types are less occurred in real-world, we still sample equal numbers for them with that of single-sentiment instances. Thus, we get a more challenging distribution than uniform sampling. Here is the details for the three types.
\begin{itemize}[leftmargin=*]
\item A \textbf{single-sentiment} instance contains one or more sentiment expression snippets, each of which includes obvious emotional words. Meanwhile, all snippets have the same sentiment polarity. For example, the instance ``\textit{It is a very disappointing photo shop with poor service attitude.}'' has two sentiment snippets: ``\textit{a very disappointing photo shop}'' and ``\textit{with poor service attitude}'', each with a negative label. 
\item A \textbf{multiple-sentiment} instance contains multiple sentiment expression snippets and more than one sentiment polarities. For example, the instance ``\textit{The packing is beautiful, but there are scratches in the disk.}'' has two sentiment snippets separated by commas, where the former is positive and the latter is negative. Multiple-sentiment instances require a model to be able to recognize the main sentiment from several sentiment expression snippets.
\item An \textbf{obscure-sentiment} instance has no obvious emotional words, e.g., the examples with the sentiment type of ``\textit{Obscure}'' in Figure \ref{fig:intro}. The sentiment is expressed in the form of metaphor, satire, etc. These instances require a model to have the ability of understanding and reasoning.
\end{itemize}

In order to improve annotation efficiency, we first assign a sentiment type for each instance based on rules, and then ask annotators to check and correct the assignment. Firstly, we use an emotional vocabulary to identify the obvious emotional words in a given instance. Then we split the instance into several snippets via punctuation and use a well-trained model to identify the sentiment of each snippet. According to the definition of each type, we automatically specify a type for each instance. At last, we ask annotators to check types of all instances and correct the error ones.

\subsection{Adversarial Instance Creation}
\label{ssec:data_robust}

After the instance collection, manual-annotated adversarial instances are created for all instances. Specifically, we first define five adversarial types, as listed in the first column of Figure \ref{fig:intro}. Then, for each original instance $t$, the annotators are required to construct a corresponding adversarial instance $t{'}$ according to adversarial types.
It is worth noting that in order to ensure the distribution of generated data to be consistent with the original one, only one adversarial instance is generated based on a certain adversarial type for each original instance.
If an original instance corresponds to multiple generated adversarial instances with different adversarial types, we retain one generated instance randomly. Besides, in order to guarantee the authenticity of the data, the generated adversarial instances are required to be not only conformed to common sense, but also to be coherent and complete.
The five adversarial types defined in our dataset are listed as follows.
\begin{itemize}[leftmargin=*]
\item \textbf{IDR of Dispensable Words}. Insert/Delete/Replace (IDR) the dispensable words which have no influence on the sentiment prediction of the whole sentence.
\item \textbf{Negative Description}. We replace the phrase in the original instance with a negative expression, such as ``dislike'' instead of ``hate''. Meanwhile, the sentiment label changes with the changes in the instance.
\item \textbf{Obscure Expression}. The sentiment expression changes from explicit to implicit, that is, the sentiment is expressed in the form of metaphor, satire, etc.
\item \textbf{Paraphrase of LF Words}. Some sentiment words are replaced by Low-Frequency (LF) words with the same semantics.
\item \textbf{Commonsense Knowledge (CSK)}. The words which are not sentiment words but are influential to the prediction are replaced. Usually, the sentiment of the adversarial instance will change, and commonsense knowledge is needed both in the replacement and the sentiment prediction.
\end{itemize}

In addition, sentiment polarity is labeled both for the original instance and the corresponding adversarial instance in this process.

\subsection{Explanation Annotation}
\label{ssec:data_inter}

Given an instance and the corresponding sentiment label, the annotators select tokens from instances to form an explanation which can support the prediction of the label. In our work, an explanation consists of several tokens and should satisfy the following three properties.
\begin{itemize}[leftmargin=*]
\item \textbf{Sufficient}. An explanation is sufficient if it contains enough information for people to make the right prediction. In other words, it is easy for people to make correct prediction just based on the tokens in the explanation.
\item \textbf{Understandable}. An explanation is understandable if it can be understood by humans. Usually, it is clear and coherent.
\item \textbf{Compact}. An explanation is compact if all of its tokens are indeed required in making a prediction. That is to say, when any token is removed from the explanation, the prediction will change or be difficult to make.
\end{itemize}

According to the above criteria, some instances may contain more than one explanation set. As shown in Figure \ref{fig:intro}, the last example contains two explanation sets, and tokens in the same color belong to the same set. In the annotation process, we ask annotators to annotate all explanations and organize them into several sets, each of which should support prediction independently. 

Our explanation annotation consists of three steps. Firstly, we ask annotators to annotate all explanation sets. Then, based on the triple of (instance, explanation, label), another annotators score each explanation set according to the above three properties. For sufficiency, annotators rate their confidences on a 3-point scale consisting of \textit{can not support label (1)}, \textit{not sure (2)} and \textit{can support label (3)}. For understandability, annotators rate their confidences on a 4-point scale including \textit{can not be understood (1)}, \textit{difficult to be understood (2)}, \textit{can be understood (3)} and \textit{easy to be understood (4)}. For compactness, annotators rate their confidences on a 4-point scale composed of \textit{including redundant tokens (1)}, \textit{including disturbances (2)}, \textit{not sure (3)} and \textit{conciseness (4)}. At last, we filter out low-quality explanations according to these scores, and ask annotators to correct these explanations. Then the corrected explanations are scored again. This process iterates until all data meet the criteria\footnote{The sufficiency score is 3, the understandability score is more than 2 and the compactness score is 4.}.

\subsection{Data Statistics}
\label{ssec:data_stas}

We summarize the statistics of our new evaluation set in Table \ref{tab:data_stas}. 
Our evaluation dataset contains 1,502 original instances from real users and the corresponding adversarial instances created by humans. The adversarial instances cover five common adversarial types, and the type of \textit{Obscure} corresponds to most instances.
The original/adversarial pairs can be used to evaluate model robustness and interpretability from the perspective of the faithfulness.
The average explanation set number is $1.4$, and the average explanation length ratio is $29.0\%$.

\begin{table*}[tb]
\renewcommand\tabcolsep{2.5pt}
\centering
\begin{tabular}{c c c | c c c c c | c c}
\toprule
\multicolumn{3}{c|}{Sentiment Type} & \multicolumn{5}{c|}{Adversarial Type} & \multicolumn{2}{c}{Manual Explanation} \\
\cline{1-3} \cline{4-8} \cline{9-10}
Single & Multiple & Obscure & IDR of Words & Negative & Obscure & LF Words & CSK & Avg LenR & Avg SetN \\
\hline
1,000 & 1,004 & 1,000 & 347 & 183 & 774 & 158 & 40 & 29.0\% & 1.4\\
\bottomrule
\end{tabular}
\caption{The Statistics of evaluation set. ``Avg LenR'' represents the average of the ratio of the explanation length to the instance length, and ``Avg SetN'' represents the average number of the explanation sets.}
\label{tab:data_stas}
\end{table*}  
\section{Evaluation Metrics}
\label{sec:metric}
In this section, we introduce the metrics for evaluating the model robustness and interpretability following the literature (\cite{wang2019towards, jacovi2020towards, deyoung2020eraser}). We evaluate robustness using the accuracy on the challenging dataset, and evaluate interpretability from the perspectives of plausibility and faithfulness based on the annotated explanations. The following is the specific metrics for these attributes. Our evaluation script will be released at \url{https://github.com/xyz}.

\textbf{Accuracy for Robustness}. We use the Accuracy on our evaluation dataset to evaluate model robustness. In order to analyze the robustness of the model in detail, we report the model accuracy on the original instances and the adversarial instances respectively.

\textbf{F1 for Plausibility}. The plausibility is to measure how well the explanation provided by models agree with that annotated by humans. As our evaluation dataset gives token-level based explanations, we take \textit{Micro-F1} and \textit{Macro-F1} as evaluation metrics. An input text may contain several explanation sets, each of which can be regarded as an independent support for the prediction. For the sake of fairness, we take the set that has the largest F1-score with the predicted explanations as the gold set for the current prediction. 

\textbf{New\_P for Faithfulness}. The faithfulness aims to reflect the true reasoning process of the model when making a decision (\cite{jacovi2020towards}). Following \citet{deyoung2020eraser}, we evaluate faithfulness in terms of sufficiency and compactness based on a single instance. \textit{Sufficiency} measures whether the explanation contains enough information for a model to make a prediction. As described in Equation \ref{equation:e2}, $F(x)_j$ represents the prediction probability provided by the model $F$ for the predicted class $j$, and $e_i$ represents the explanation of instance $x_i$. A low score here implies the explanation is indeed sufficient in the prediction. \textit{Compactness} measures whether the explanation only contains tokens that are indeed influential in the prediction, as shown in Equation \ref{equation:e3}, where ($x_i\setminus{e_i}$) represents the explanation $e_i$ is removed from the instance $x_i$. A high score means the explanation is indeed influential in the prediction.
\begin{equation}
\small
\centering
Score_{suf} = F(x_i)_j - F(e_i)_j
\label{equation:e2}
\end{equation}
\begin{equation}
\small
\centering
Score_{com} = F(x_i)_j - F(x_i \setminus e_i)_j
\label{equation:e3}
\end{equation}
In our work, we define a new precision $New\_P$ to integrate $Score_{suf}$ and $Score_{com}$, as shown in Equation \ref{equation:e4}.   
\begin{equation}
\small
\centering
new\_F(x_i) = \left\{
\begin{array}{rcl}
1 &  & y_{e_i} = y_{x_i}, y_{x_i\setminus e_i} \ne y_{x_i} \\
0 &  & others \\
\end{array}\right\}
\label{equation:e4}
\end{equation}
Where $y_{x}$ represents the output class predicted by the model $F$ based on the input $x$.

\textbf{MAP for Faithfulness}. In addition, we propose to evaluate faithfulness based on the consistency of explanations under perturbations. According to the pair of original instance and its corresponding adversarial instance, we use Mean Average Precision (MAP) to evaluate the consistency of their token importance lists, as shown below.
\begin{equation}
\small
\centering
MAP = \frac{\sum_{i=1}^{|X^a|}(\sum_{j=1}^i F(x_j^a, X_{1:i}^o))/i}{|X^a|}
\label{equation:e1}
\end{equation}
Where $X^o$ and $X^a$ represent the sorted token importance list of the original instance and the adversarial instance, respectively. $|X^a|$ represents the number of tokens in the list $X^a$. $X_{1:i}^o$ represents a portion of the list $X^o$, which consists of top-$i$ tokens. The function $F(x, Y)$ is used to determine whether the token $x$ belongs to the list $Y$. If $x$ is in the list $Y$, $F(x, Y)$ returns 1. The high MAP indicates the high consistency.


\section{Base Models}
\label{sec:models}
We focus on the trustworthiness evaluation itself, rather than on any particular model or interpretation method. In this work, we use three popular post-hoc interpretation methods to evaluate four baseline models with different network complexity.

\subsection{Evaluated Models}
\label{ssec:test_model}
To find the relationship between accuracy and trustworthiness, we implement four baseline models with different network architectures and parameter sizes, namely LSTM (\cite{hochreiter1997long}), ERNIE-base/large (\cite{sun2019ernie}) and SKEP (\cite{tian2020skep}). 
All models are trained on the ChnSentiCorp\footnote{\url{http://www.searchforum.org.cn/tansongbo/corpus-sen-ti.htm}} dataset and implemented based on PaddleNLP\footnote{\url{https://github.com/PaddlePaddle/PaddleNLP}}. We use default values for all hyper-parameters.

\textbf{LSTM}. We take the BiLSTM-based model as the baseline model with low complexity.

\textbf{ERNIE}. Considering the model complexity from the perspective of network architecture, we take transformer-based pre-trained models as baseline models with high complexity. In our experiments, we use ERNIE as the vanilla pre-trained model. In order to evaluate models with different parameter sizes, we implement ERNIE-base and ERNIE-large. 

\textbf{SKEP}. SKEP incorporates sentiment knowledge into vanilla pre-trained models to enhance representation and achieves SOTA results on most benchmark datasets of the sentiment analysis task. Compared with vanilla pre-trained models, SKEP utilize task-specific knowledge to enhance model capability. In order to verify the impact of domain knowledge on the trustworthiness of the model, we take SKEP as a baseline model.

The performance of these four models on the ChnSentiCorp test set is shown in Table \ref{tab:all_result}.

\subsection{Interpretation Methods}
\label{ssec:inter_model}
We implement three types of interpretation methods, i.e., gradient-based, attention-based and linear-based, to extract explanations from a well-trained model. The source codes of these methods will be released at \url{https://github/xyz}.

%
\textbf{Integrated Gradient (IG) Method}. \citet{sundararajan2017axiomatic} introduce IG method, an improved version of vanilla gradient-based method which assigns importance score for each token by the loss gradient (\cite{simonyan2014deep}). They define a baseline $x_0$, which is an input absent of information. Token importance is determined by integrating the gradient along the path from this baseline to the original input. In the experiments, we use a sequence of all zero embeddings as the baseline $x_0$ for each original input. And we set step size to 300.


\textbf{Attention-based (ATT) Method}. The methods of this kind take attention scores (\cite{bahdanau2014neural}) as token importance scores. For the model fine-tuned by a pre-trained model, attention scores are taken as the self-attention weights induced from the [CLS] token index to all other indices in the last layer. As the pre-trained model uses wordpiece tokenization, in order to compute a score for a token, we sum the self-attention weights assigned to its constituent pieces. Meanwhile, the pre-trained model is also multi-headed, so we average scores over heads to derive a final score.

\textbf{LIME}. \citet{ribeiro2016should} propose a system to explain why a classifier makes a prediction by identifying useful tokens of the input. They use a linear model $g$ as the interpretation model to approximate the evaluated model locally. And they use the weighted square loss and a set of perturbed samples which contains $K$ tokens of the original input to optimize the selection of useful tokens. We set sample size to 5000 and set $K$ to 10 in our experiments. 

\begin{table*}[tb]
\renewcommand\tabcolsep{2.5pt}
\centering
\begin{tabular}{l | l | l l l | l l l | l | l l l | l l l | l l l}
\toprule
\multirow{3}{*}{Models} & \multirow{3}{*}{Acc$_{chn}$} & \multicolumn{6}{c|}{Robustness} & \multicolumn{4}{c|}{Faithfulness} & \multicolumn{6}{c}{Plausibility}\\
\cline{3-8} \cline{9-12} \cline{13-18}
 & & \multicolumn{3}{c|}{Acc} & \multicolumn{3}{c|}{Acc} & \multirow{2}{*}{MAP} & \multicolumn{3}{c|}{New\_P} & \multicolumn{3}{c|}{Macro-F1} & \multicolumn{3}{c}{Micro-F1} \\
 \cline{3-5} \cline{6-8} \cline{10-12} \cline{13-15} \cline{16-18}
 & & Ori & Adv & All & Sin & Mul & Obs & & Ori & Adv & All & Ori & Adv & All & Ori & Adv & All\\
\hline
LSTM  & 86.8 & 62.0 & 52.1 & 57.1 & 85.0 & 40.2 & 60.8 & 59.4 & 26.0 & 18.6 & 22.3 & 37.4 & 36.8 & 37.2 & 38.4 & 37.4 & 37.9 \\
ERNIE-base & 95.4 & 67.7 & 62.7 & 65.2 & 92.0 & 31.3 & 72.4 & 46.7 & 37.0 & 32.9 & 35.0 & 38.6 & 36.8 & 37.8 & 39.6 & 37.7 & 38.7 \\
ERNIE-large & 95.8 & 69.6 & 66.8 & 68.2 & 93.1 & 33.4 & 78.1 & 42.3 & 40.3 & 38.6 & 39.4 & 40.9 & 38.7 & 39.8 & 41.7 & 39.2 & 40.4\\
SKEP & 96.3 & 68.4 & 64.7 & 66.5 & 92.7 & 31.8 & 76.2 & 43.8 & 35.2 & 34.4 & 34.8 & 39.8 & 38.7 & 39.3 & 40.5 & 39.5 & 40.1\\
\bottomrule
\end{tabular}
\caption{Performances of four models on ChnSentiCorp test set and our new evaluation dataset. Acc$_{chn}$ shows the accuracy of each model on the test set of ChnSentiCorp. \textit{Robustness} describes the accuracy performance on our evaluation set. We report performance on the original instances (\textit{Ori}), manually annotated adversarial instances (\textit{Adv}) and all instances (\textit{All}). In addition, we report results on different sentiment types, including single-sentiment (Sin), multiple-sentiment (Mul) and Obscure-sentiment (Obs). \textit{Faithfulness} and \textit{Plausibility} are two criteria for interpretability evaluation. The corresponding results are obtained by the interpretation method of LIME. Similarly, we evaluate interpretability on the original instances and adversarial instances, respectively.}
\label{tab:all_result}
\end{table*}  
\section{Evaluation}
\label{sec:result}

Here we present performance of four evaluated models discussed in Section \ref{ssec:test_model}, with respect to the metrics proposed in Section \ref{sec:metric}. We first report the main results in Section \ref{ssec:main_result}. Then we show the detailed performance of robustness and interpretability in Section \ref{ssec:eval_robust} and \ref{ssec:eval_inter}, respectively.
Finally, we analyze the relationship between model accuracy and trustworthiness in Section \ref{ssec:eval_relation}.

\subsection{Main Results}
\label{ssec:main_result}
According to the performance on the dev set, we choose the best model of each evaluated method for evaluating the trustworthiness. The results are shown in Table \ref{tab:all_result}. Firstly, we can see that all models achieve promising accuracy on the test set of ChnSentiCorp, but get low performance on our new evaluation set, as illustrated by the accuracy of \textit{Ori} in the column of \textit{Robustness}. Secondly, for each model, the performance on adversarial instances is lower than that on the original instances, especially for the LSTM model. Thirdly, for each model, the performance on multiple-sentiment instances is the worst. We suspect the reason is two-fold. First, all models have defects in domain generalization and tolerance of adversarial noises, which are the common defects of NN models. Second, the pre-trained based models such as ERNIE-base, ERNIE-large and SKEP have a stronger generalization capability. The more detailed analysis is given in Section \ref{ssec:eval_robust}.

Since the LIME interpretation method is independent of the model architecture, the performance of model interpretability shown in Table \ref{tab:all_result} is evaluated based on the explanations extracted by LIME (performance of all interpretation methods is shown in Table \ref{tab:eval_inter}). It is can be seen that the plausibility of explanations is related to model accuracy, and the higher accuracy corresponds to a higher plausibility. But there is a big gap between accuracy and plausibility. 

According to \textit{MAP} results on faithfulness in Table \ref{tab:all_result}, we can see that \textit{MAP} metric is relevant to model complexity, including network architecture and parameter size. The more complexity the model, the lower the MAP. The comparison between models with the same architecture and parameter size, i.e., ERNIE-large and SKEP, shows that the learning objectives proposed in SKEP can improve the model faithfulness. Based on \textit{New\_P} results, which evaluates the explanation from the perspective of sufficiency and compactness, we can see that it is mainly positively related to plausibility.

\subsection{Robustness Evaluation}
\label{ssec:eval_robust}

\textbf{Performance on Different Adversarial
Types}. As described in subsection 3.2, we define five adversarial types in order to make a comprehensive assessment for model robustness. The particular-Acc for each adversarial type and the Micro-Acc for all types are listed in Table \ref{tab:eval_adv_type}. 
From the comparison between LSTM and other three pre-trained models, we can see that pre-trained models have a stronger generalization capability, especially on adversarial types of ``negative description'' and ``paraphrase of low-frequency words''. Moreover, large-size models such as ERNIE-large and SKEP outperform base-size models on most of types. This is also due to the stronger representation ability of large models.

From the comparison between models with the same architecture and parameter size, i.e., ERNIE-large and SKEP, we find that ERNIE-large performs better in Micro-Acc and SKEP outperforms ERNIE-large on severel adversarial types, such as ``CSK'' and ``LF Words''. We suspect the reason is three-fold. Firstly, the generated data distribution of different adversarial types is unbalanced, as shown in Table \ref{tab:data_stas}. This leads to ERNIE-large's overall performance on all types (as shown in Micro-Acc) better than SKEP. Secondly, since SKEP enhances the vanilla pre-trained model with sentiment knowledge, such as sentiment words, aspect/polarity pairs, the corresponding words are more important in model prediction. Therefore, SKEP has better performance on types of ``CSK'' and ``LF Words'', which depends on the importance of aspects in prediction. Meanwhile, SKEP has lower performance on ``Obscure'' whose instances have no obvious sentiment words. At last, SKEP is post trained on the large-scale subjectivity corpus and enhanced with sentiment knowledge extracted from the corpus, so it is much more easier to over fit the training data.

\begin{table*}[tb]
\renewcommand\tabcolsep{2.5pt}
\centering
\begin{tabular}{l | c | c | c | c | c | c }
\toprule
Models & IDR of Words & Negative & CSK & LF Words & Obscure & Micro-Acc \\
\hline
LSTM & 60.5 & 40.4 & 40.0 & 58.9 & 50.4 & 52.1\\
ERNIE-base & 71.8 & 79.4 & 58.3 & 88.1 & 54.9 & 62.7\\
ERNIE-large & \textbf{74.2} & \textbf{84.9} & 57.5 & 88.0 & \textbf{58.4} & \textbf{66.8}\\
SKEP & \textbf{74.2} & 82.1 & \textbf{59.6} & \textbf{89.3} & 55.1 & 64.7 \\
\bottomrule
\end{tabular}
\caption{Performance on adversarial instances with different adversarial types.}
\label{tab:eval_adv_type}
\end{table*} 

\begin{table*}[tb]
\renewcommand\tabcolsep{2.5pt}
\centering
\begin{tabular}{l | c | c | c}
\toprule
\multirow{2}{*}{Models} & Original Instances & Pattern-based Instances (TextFlint) & Manual Adversarial Instances \\
& (1,000) & (4,878) & (1,000) \\
\hline
SKEP & 91.4  & 85.9 & 76.8 \\
SKEP + Augmented data & 89.7 (\textbf{-1.7})  & 91.3 (\textbf{+5.4}) & 74.0 (\textbf{-2.8}) \\
\bottomrule
\end{tabular}
\caption{Performance on different adversarial generation methods.} 
\label{tab:eval_adv_generation}
\end{table*} 

\textbf{Comparison between adversarial generation methods}. Previous works (\cite{ribeiro2020beyond,gui2021textflint}) exploit fallible patterns summarized from the dataset to generate adversarial examples for model robustness evaluation. However, lots of unnatural examples which don't occur in real applications are generated. In order to verify the effect of automatically generated data on model robustness evaluation in real applications, we conduct experiments on adversarial instances with different generation methods. Firstly, we construct three evaluation test set, as shown in Table \ref{tab:eval_adv_generation}. From our challenging evaluation dataset, we select instances in types of single-sentiment and obscure-sentiment to construct the ``\textit{Original instances}''\footnote{Since it is difficult or even impossible for pattern-based methods (such as TextFlint used in our experiments) to generate high-quality adversarial instances for multiple-sentiment type. For the sake of fairness, we only select instances of single-sentiment and obscure-sentiment types.}. And the corresponding adversarial instances of them are used to construct the ``\textit{Manual Adversarial Instances}''. Both of the two set contain 1,000 instances. Meanwhile, based on the selected original instances, we use TextFlint (\cite{gui2021textflint}) to generate 4,878 adversarial instances for them, denoted as ``Pattern-based Instances (TextFlint)'' in Table \ref{tab:eval_adv_generation}. Then we use the SKEP model described in Section \ref{ssec:test_model} as the baseline model. And it is enhanced with augmented data generated by TextFlint based on the ChnSentiCorp training data, denoted as ``SKEP + Augmented data'' in Table \ref{tab:eval_adv_generation}.

From experiment results in Table \ref{tab:eval_adv_generation}, we find that SKEP enhanced by data augmentation improves performance on pattern-based instances by 5.4\%. However, it does not perform well on the selected original instances and the corresponding manually annotated adversarial instances, decreasing their accuracy by 1.7 and 2.8 respectively. It can be see that data generated by pattern-based methods can improve performance on data of the same distribution, but it is ineffective on robustness evaluation in real applications. Therefore, manually annotated data is very necessary for model robustness evaluation.

\subsection{Interpretability Evaluation}
\label{ssec:eval_inter}

\begin{table*}[tb]
\renewcommand\tabcolsep{2.5pt}
\centering
\begin{tabular}{l | l | l l l | l l l | l l l}
\toprule
\multirow{3}{*}{Models + Methods} & \multicolumn{4}{c|}{Faithfulness} & \multicolumn{6}{c}{Plausibility}\\
\cline{2-5} \cline{6-11}
 & \multirow{2}{*}{MAP} & \multicolumn{3}{c|}{New\_P} & \multicolumn{3}{c|}{Macro-F1} & \multicolumn{3}{c}{Micro-F1} \\
\cline{3-5} \cline{6-8} \cline{9-11}
 &  & Ori & Adv & All & Ori & Adv & All & Ori & Adv & All\\
\hline
LSTM + IG & 59.6 & 29.6 & 22.3 & 26.0 & 39.1 & 38.3 & 38.8 & 39.9 & 38.7 & 39.3\\
LSTM + ATT & 69.8 & 21.1 & 22.4 & 21.7 & 33.6 & 32.8 & 33.3 & 36.7 & 35.3 & 36.0\\
LSTM+ LIME & 59.4 & 26.0 & 18.6 & 22.3 & 37.4 & 36.8 & 37.2 & 38.4 & 37.4 & 37.9\\
\hline
ERNIE-base + IG & 36.4 & 26.0 & 26.5 & 26.2 & 35.1 & 35.2 & 35.1 & 36.4 & 36.2 & 36.3\\
ERNIE-base + ATT & 65.1 & 18.2 & 19.5 & 18.8 & 24.3 & 24.8 & 24.6 & 27.9 & 27.7 & 27.8\\
ERNIE-base + LIME & 46.7 & 37.0 & 32.9 & 35.0 & 38.6 & 36.8 & 37.8 & 39.6 & 37.7 & 38.7\\
\hline
ERNIE-large + IG & 34.1 & 24.2 & 24.0 & 24.1 & 38.2 & 37.4 & 37.8 & 39.0 & 37.9 & 38.5\\
ERNIE-large + ATT & 64.6 & 17.5 & 17.9 & 17.7 & 28.0 & 27.7 & 27.9 & 31.0 & 30.3 & 30.7\\
ERNIE-large + LIME & 42.3 & 40.3 & 38.6 & 39.4 & 40.9 & 38.7 & 39.8 & 41.7 & 39.2 & 40.4\\
\hline
SKEP + IG & 30.5 & 24.2 & 24.8 & 24.5 & 38.7 & 38.6 & 38.6 & 39.3 & 38.8 & 39.0\\
SKEP + ATT & 60.7 & 20.3 & 22.2 & 21.2 & 29.8 & 29.5 & 29.6 & 33.1 & 32.1 & 32.5\\
SKEP + LIME & 43.8 & 35.2 & 34.4 & 34.8 & 39.8 & 38.7 & 39.3 & 40.5 & 39.5 & 40.1\\
\bottomrule
\end{tabular}
\caption{Performance on interpretability, from the perspectives of faithfulness and plausibility.}
\label{tab:eval_inter}
\end{table*} 

For each evaluated model, we use different interpretation methods to assign importance score to each token. We take the top-$k$ tokens as explanations for each instance, where $k$ is obtained by multiplying the average length ratio of explanations (see Table \ref{tab:data_stas}) and the length of the instance. Then we use the extracted explanations to evaluate the interpretability of the model from the perspectives of faithfulness and plausibility. Table \ref{tab:eval_inter} shows performance of different evaluated models with different interpretation methods. Next, we will analyze the two criteria in detail.

\textbf{Faithfulness}. Firstly, we compare the faithfulness performance of different evaluated models.
According to MAP which evaluates the consistency of explanations under perturbations, LSTM model has the highest faithfulness regardless of different interpretation methods. Similarly, ERNIE-base has higher faithfulness than ERNIE-large based on any listed interpretation method. It can be seen that model faithfulness mainly depends on model architecture, and the simpler the architecture, the higher the faithfulness. We suspect that models with complex architectures are more sensitive, resulting in lower faithfulness.
From the comparison between ERNIE-large and SKEP, we find that SKEP can improve the model faithfulness based on some specific interpretation methods, such as LIME. This may be caused by SKEP incorporating sentiment knowledge into pre-trained models.

When comparing different interpretation methods, we observe that ATT method outperforms other interpretation methods across all evaluated models, which also confirms the viewpoint in \cite{wiegreffe2019attention}.
Another trend worth noting is that the faithfulness ranking of the three interpretation methods is the same with different pre-trained models, i.e., in the order of ATT, LIME, IG. This again shows that faithfulness is related to model architecture.

\textbf{Plausibility}. From Table \ref{tab:eval_inter}, we observe that LIME consistently performs better than other interpretation methods regardless of different evaluated models, and ATT method has the worst performance. Since LIME is model-agnostic, the plausibility of its explanations is positively correlated with model accuracy (see Table \ref{tab:all_result}). However, there is a big gap between model accuracy and plausibility. Other interpretation methods are related to model architecture, e.g., the plausibility of LSTM is better than transformer-based models. For transformer-based models, the plausibility mainly depends on model accuracy, and the ranking of the three interpretation methods is mainly the same, i.e., in the order of LIME, IG, and ATT, which is roughly the same trend as the faithfulness results. From the comparison between ERNIE-large and SKEP, we observe that SKEP outperforms ERNIE-large with IG and ATT methods. We suspect this is because incorporating sentiment knowledge into pre-trained models can enhance representations of specific tokens and lead to the variation of token importance score.

Finally, based on the results in Table \ref{tab:eval_inter}, it can be seen that the plausibility of model is mostly positively correlated with model accuracy, especially for models with the same architecture.

\begin{figure}[tb]
\centering
\includegraphics[width=0.95\textwidth]{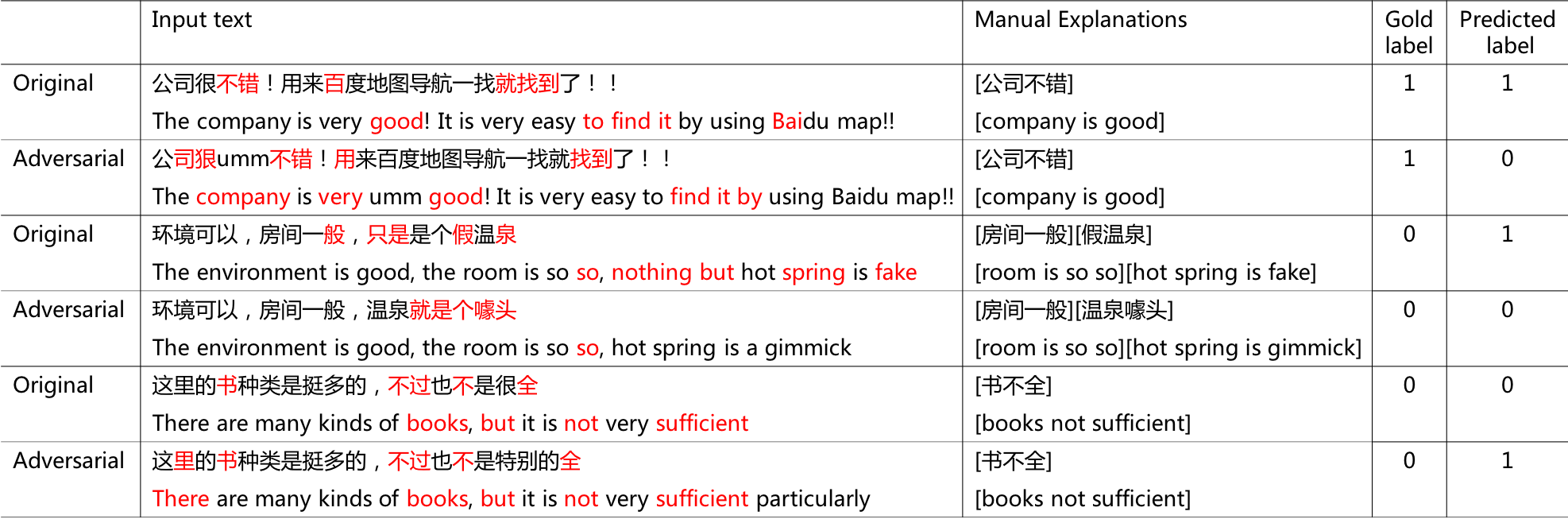}
\caption{Examples with ERNIE-large model and LIME interpretation method. The bold red tokens in the input text belong to top $k$ tokens by LIME and constitute the explanation.}
\label{fig:case_study}
\end{figure}

\textbf{Ablation study}. From Figure \ref{fig:case_study}, we can see that ERNIE-large is context-sensitive. The disturbances in the input text can lead to changes in the distribution of token importance. Moreover, slight changes in explanations may lead to error predictions, as shown in the last two examples. How to tolerate disturbances is a long-time problem that needs to be overcome by the model.

\subsection{Relationship Between Accuracy and Trustworthiness}
\label{ssec:eval_relation}
We have analyzed performance of robustness and interpretability in detail. Then we discuss the relationship between accuracy and trustworthiness.
Robustness is related to the generalization ability of the model. The Model fine-tuned on pre-trained models has better robustness, as the representations learned by pre-trained models have been enhanced by the large scale corpus. But no single `off-the-shelf' model can perform well on instances with different adversarial types, as shown in Table \ref{tab:eval_adv_type}.
The faithfulness is mainly connected with model complexity, including network architecture and parameter size. Based on the same complexity, the faithfulness can be improved by model optimization, such as SKEP. 
The plausibility is mainly relevant to model accuracy, especially for models with the same architecture. There are two points worth noting: 1) plausibility is very low compared with accuracy; 2) Based on ATT and gradient-based interpretation methods, LSTM model has high plausibility, but its accuracy is low. How to utilize explanations to improve accuracy is a problem to be resolved.
\section{Conclusion}
\label{sec:conclusion}

In order to comprehensively evaluate the trustworthiness of deep learning models, we propose a well-annotated sentiment analysis dataset, which contains three types of manual annotations, including challenging distribution of instances, manual adversarial instances and sentiment explanations. This data facilitates us to systematically study the factors for building trustworthy AI systems. 
We implement four typical models and four evaluation metrics, and analysis the relation between accuracy, robustness and interpretability using this data. Experiment results show that the necessity and usefulness of manual annotated trustworthy dataset. Firstly, despite many deep learning models perform well in terms of accuracy, their robustness and interpretability are far from satisfaction. Secondly, automatically generated data has limited effect on evaluating models in real-applications. Thirdly, larger models are often too sensitive to the small change of context, resulting in a bad faithfulness performance for interpretability. In the future, we would like to extend our work to more NLP tasks. We hope it can facilitate progress on several directions, such as better evaluation metrics for interpretability, causal analysis of NLP models.

\bibliographystyle{unsrtnat}
\bibliography{template}  

\end{document}